\def\year{2021}\relax
\documentclass[letterpaper]{article} 
\usepackage{aaai21}  
\usepackage{times}  
\usepackage{helvet} 
\usepackage{courier}  
\usepackage[hyphens]{url}  
\usepackage{graphicx} 
\usepackage{natbib}  
\usepackage{caption} 

\usepackage{booktabs}       
\usepackage{nicefrac}       
\usepackage{xspace}
\usepackage{amssymb}
\usepackage{multirow}
\usepackage{subfig}
\usepackage{amsmath}
\usepackage[prependcaption]{todonotes}
\usepackage{graphicx,wrapfig,lipsum}
\usepackage{subfig}
\usepackage{xcolor}
\usepackage{fancyvrb}
\usepackage[export]{adjustbox}
\usepackage{colortbl}
\usepackage{array}
\usepackage{bm}
\usepackage{booktabs,arydshln}

\title{Measuring Societal Biases from Text Corpora with Smoothed First-Order Co-occurrence}

\author{
      Navid Rekabsaz,\textsuperscript{\rm 1}
      Robert West,\textsuperscript{\rm 2}
      James Henderson,\textsuperscript{\rm 3}
      Allan Hanbury\textsuperscript{\rm 4}\\
}

\affiliations {
\textsuperscript{\rm 1}Johannes Kepler University, Linz Institute of Technology, AI Lab, \\
\textsuperscript{\rm 2}EPFL, \\
\textsuperscript{\rm 3}Idiap Research Institute, \\
\textsuperscript{\rm 4}TU Wien, Complexity Science Hub Vienna \\
navid.rekabsaz@jku.at,
robert.west@epfl.ch,
james.henderson@idiap.ch,
allan.hanbury@tuwien.ac.at\\
}

\newcolumntype{L}[1]{>{\raggedright\let\newline\\\arraybackslash\hspace{0pt}}m{#1}}
\newcolumntype{C}[1]{>{\centering\let\newline\\\arraybackslash\hspace{0pt}}m{#1}}
\newcolumntype{R}[1]{>{\raggedleft\let\newline\\\arraybackslash\hspace{0pt}}m{#1}}

\makeatletter
\def\adl@drawiv#1#2#3{%
        \hskip.5\tabcolsep
        \xleaders#3{#2.5\@tempdimb #1{1}#2.5\@tempdimb}%
                #2\z@ plus1fil minus1fil\relax
        \hskip.5\tabcolsep}
\newcommand{\cdashlinelr}[1]{%
  \noalign{\vskip\aboverulesep
           \global\let\@dashdrawstore\adl@draw
           \global\let\adl@draw\adl@drawiv}
  \cdashline{#1}
  \noalign{\global\let\adl@draw\@dashdrawstore
           \vskip\belowrulesep}}
\makeatother

\def\eqref#1{equation~\ref{#1}}

\def\1{\bm{1}}

\def\ve{{\bm{e}}}

\def\vu{{\bm{u}}}
\def\vv{{\bm{v}}}

\def\mD{{\bm{D}}}

\def\mU{{\bm{U}}}
\def\mV{{\bm{V}}}

\DeclareMathAlphabet{\mathsfit}{\encodingdefault}{\sfdefault}{m}{sl}
\SetMathAlphabet{\mathsfit}{bold}{\encodingdefault}{\sfdefault}{bx}{n}

\def\sP{{\mathbb{P}}}

\def\sR{{\mathbb{R}}}

\def\sV{{\mathbb{V}}}

\newcommand{\biasPC}{\textsc{Directional}\xspace}
\newcommand{\biascentroid}{\textsc{Centroid}\xspace}
\newcommand{\biasmeanSim}{\textsc{Average}_{\textsc{High}}\xspace}
\newcommand{\biasmeanCo}{\textsc{Average}_{\textsc{First}}\xspace}

\def\bias{{\psi}}

\newcommand{\hisg}{\text{eSG}\xspace}
\newcommand{\higlove}{\text{eGloVe}\xspace}
\newcommand{\initglove}{\text{initGlove}\xspace}

\definecolor{darkfemale}{HTML}{e66101}
\definecolor{darkmale}{HTML}{5e3c99}
\definecolor{lightfemale}{HTML}{fdb863}
\definecolor{lightmale}{HTML}{b2abd2}
\definecolor{gray}{HTML}{bdbdbd}
\definecolor{cadmiumgreen}{rgb}{0.0, 0.42, 0.24}

\begin{document}

\maketitle

\begin{abstract}
Text corpora are widely used resources for measuring societal biases and stereotypes. The common approach to measuring such biases using a corpus is by calculating the similarities between the embedding vector of a word (like \emph{nurse}) and the vectors of the representative words of the concepts of interest (such as genders). In this study, we show that, depending on what one aims to quantify as bias, this commonly-used approach can introduce non-relevant concepts into bias measurement. We propose an alternative approach to bias measurement utilizing the smoothed first-order co-occurrence relations between the word and the representative concept words, which we derive by reconstructing the co-occurrence estimates inherent in word embedding models. We compare these approaches by conducting several experiments on the scenario of measuring gender bias of occupational words, according to an English Wikipedia corpus. Our experiments show higher correlations of the measured gender bias with the actual gender bias statistics of the U.S. job market -- on two collections and with a variety of word embedding models -- using the first-order approach in comparison with the vector similarity-based approaches. The first-order approach also suggests a more severe bias towards female in a few specific occupations than the other approaches.

\end{abstract}

\section{Introduction}
Text data has been widely utilized for studying and monitoring societal phenomena -- such as biases and stereotypes -- commonly by exploiting co-occurrence statistics of words in text. In these approaches, a societal bias construct (an unobservable abstraction that we aim to characterize) is quantified using measures of words association. A word such as \emph{nurse} is considered to be stereotypically biased towards the female concept, when a significant imbalance is observed between the associations of \emph{nurse} to female versus male concept. Each of the concepts is commonly defined by a group of words, referred to as \emph{concept-representative words}. The focus of the present work is on the computational methods for measuring biases from text corpora -- a particularly essential component in various social studies.

The common approach to calculate words associations for bias measurement is by adopting word embedding models trained on text corpora as in preceding studies~\cite{lenton2009latent,hoyle2019unsupervised,zhou2019examining,chang2019automatically,zhao2019gender,garg2018word,caliskan2017semantics,bolukbasi2016man}. In these studies, the associations of words to concepts are measured based on some form of vector similarity, for instance by using the cosine metric. The present study sheds light on and discusses what is captured as bias by these vector similarity-based approaches, and proposes a complementary bias measurement approach based on a smoothed variant of direct (first-order) co-occurrences. Let us first have a closer look at what is measured by vector similarity or more generally by similarity metrics applied to distributional representations.

\begin{figure}[t]
  \includegraphics[width=0.4\textwidth,center]{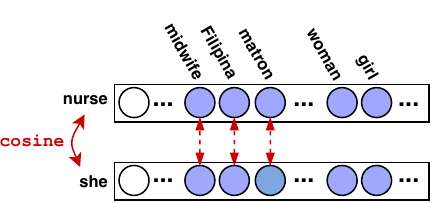}
  \caption{Word vectors with explicit dimensions. Estimating bias with similarity between the vectors also counts some (arguably) irrelevant context-words to the female concept (red dashed lines). The example is based on the results in Table~\ref{tbl:genderbias:occpations_topdims}.}
  \label{fig:introduction:circular}
\end{figure}

In distributional representations, two vectors are more similar if the corresponding words both frequently co-occur with a set of context-words (second-order co-occurrence). Figure~\ref{fig:introduction:circular} elaborates this using a toy example. In the example, the association of the word \emph{nurse} to the female concept, represented by the word \emph{she}, is calculated using cosine similarity between their vectors. The word vectors in the example are high-dimensional representations, such that each dimension of a vector is defined explicitly with a specific context-word, and each value of the vector represents the first-order co-occurrence relation between the word and the corresponding context-word. We refer to such vectors as \emph{explicit representations}. As shown, \emph{nurse} and \emph{she} are similar since they co-occur with several common context-words, depicted with the blue circles in both vectors. 

Let us assume that in this example our objective of gender bias measurement can be formulated as ``to quantify the extent to which \emph{nurse} is perceived as female versus male''. Considering this objective, we observe that the \emph{nurse}-to-\emph{she} similarity is influenced by the common context-words such as \emph{woman} and \emph{girl}, which are typically considered as good representatives of the female concept~\cite{garg2018word,bolukbasi2016man}. However, this similarity is also affected by several other context-words, such as \emph{midwife} and \emph{matron}. According to the literal definitions of these words (as defined in a dictionary), \emph{midwife} is gender-neutral, and \emph{matron} is a mixture of the female concept and the concept of ``being in charge of medical arrangements''. Considering this, one can argue that such common context-words can introduce partially or completely irrelevant concepts to female into the measured association, and hence into the calculated gender bias.

While we discuss this issue on explicit representations, it is also present in low-dimensional embeddings, although such an explicit dividing of dimensions into groups is not per se possible. Another difference between explicit representations and embeddings is that, since word embeddings are defined in low dimensions, the similarity of embedding vectors does not only capture second-order co-occurrences, but other orders of co-occurrence, such as first-order as well as higher orders.\footnote{For instance, \citet{kontostathis2006framework} found that Latent Semantic Analysis~\cite{deerwester1990indexing} can take into account up to fifth-order co-occurrences.} We therefore refer to the approaches that use word embeddings for quantifying bias as \emph{high-order bias measurements}.

We approach the discussed issue in high-order bias measurement by revisiting the utilization of first-order co-occurrences for measuring bias. Our proposed approach, referred to as \emph{first-order bias measurement}, estimates the association of a concept and a word by averaging the first-order co-occurrences between the word and the concept-representative context-words. The first-order bias measurement introduces an alternative to high-order approach, and has the advantage of only taking into account the context-words which are strongly related to the concept of interest.

First-order co-occurrence of words has been widely used to calculate societal phenomena particularly through counting and weighting words~\cite{monroe2008fightin,kirchler1992adorable,rekabsaz2017volatility}. Among various metrics, the ones based on Pointwise Mutual Information (PMI)~\cite{church1990word} are commonly used to measure words co-occurrence in local contexts. As mentioned by \citet{lenton2009latent}, a draw back of such count-based co-occurrence metrics is the high sparsity of their resulting vectors, as many related words never appear in the same local context. 

We address the issue of sparsity in count-based metrics by proposing two novel explicit representations, created from pre-trained word2vec Skip-Gram~\cite{mikolov2013distributed}, and GloVe~\cite{pennington2014glove} models. The proposed explicit variants exploit the word and context embedding vectors of word2vec and GloVe to estimate forms of the co-occurrence relations. Such co-occurrence relations, achieved from the reconstruction of explicit vectors from low-dimensional embeddings, provide smoothed variants of the count-based co-occurrence estimations.

We use the discussed word representations, trained on an English Wikipedia corpus to study the characteristics of the first- and high-order bias measurement approaches. We conduct several experiments on the gender bias of occupations. We first revisit the experiments conducted in previous studies~\cite{garg2018word,caliskan2017semantics} on the correlations of the gender bias of some occupations, measured using the discussed methods, to the actual statistics of gender bias in the U.S. job market. We use two collections, provided by \citet{zhao2018gender} and \citet{garg2018word}. We observe that, in all studied word representation models and the two collections, the results of our proposed first-order bias measurement shows higher correlations in comparison with the high-order approaches. 

Next, we analyze the measured gender bias of around 500 occupations using first- and high-order approaches, observing several cases of the influence of non-relevant context-words on the results of the high-order bias measurement method. Overall, our results suggest the existence of a more severe degree of bias towards female in the underlying corpus, previously undetected by high-order approaches. 

Finally, we study how each bias measurement approach reacts to (hypothetical) changes in the corpus, particularly when the corpus moves towards a more balanced representation of genders. To this end, we manipulate the corpus using the Counterfactual Data Augmentation (CDA) method~\cite{zhao2018gender,lu2018gender,zmigrod2019counterfactual}, such that the genders are represented in the augmented corpora in a more balanced way. Our observations show that, while both bias measurement methods report a decrease of gender bias, the first-order bias measurement is more sensitive (reacts faster) to the changes in the corpus. 

\emph{Limitations of the study.} Gender is treated in this study as a binary construct, and the definition of gender bias is limited to the disparity between female and male. We acknowledge that this choice neglects the broad meaning of gender, but the decision is necessary for taking practical steps. Our study is also limited to the English language. Our introduced method is however generic and can be applied to other languages as well as other forms of societal biases such as related to race, age and ethnicity.

\emph{Outline of the paper.} We first discussed related work, followed by the relevant previous methods. Our bias measurement approach is introduced next. Finally, the gender bias experiments are described, and whose results are presented.




\section{Related Work}

Various aspects of word embeddings and distributional representations in areas such as social sciences and psychology are studied in previous work. \citet{lenton2009latent} discuss the use of Latent Semantic Analysis~\cite{deerwester1990indexing} for measuring gender bias, highlighting the sparsity issue of the first-order method based on count-based metrics. The present work complements this study by exploring the benefits of a smoothed first-order bias measurement approach as an alternative to previous approaches. In a recent study, \citet{gunther2019vector} discuss the applications and common misconceptions of distributional semantic models in psychology.

Several pieces of work exploit word embeddings to study societal aspects. \citet{garg2018word} investigate the changes in gender- and race-related stereotypes over decades using historical text data. \citet{caliskan2017semantics} and more recently \citet{chang2019automatically} study the patterns of language use, indicating accurate imprints of historical biases. \citet{bolukbasi2016man} show the reflection of gender stereotypes in word analogies derived from word embeddings. \citet{zhou2019examining} propose methods to measure societal biases in languages with grammatical gender. Our work directly contributes to these studies by proposing a more accurate approach for measuring bias in corpora. In this line, \citet{hoyle2019unsupervised} propose a method to measure the differences between descriptions of men and women. In contrast to our bag-of-words-based method, they measure bias using a parsed corpus.

Gender bias is also studied in various downstream tasks, such as sentiment analysis~\cite{kiritchenko2018examining}, visual semantic role labeling~\cite{zhao2017men}, coreference resolution~\cite{zhao2018gender,rudinger2018gender}, information retrieval~\cite{rekabsaz2020neural} text, and classification~\cite{dixon2018measuring,barrett2019adversarial,elazar2018adversarial,de2019bias}, as well as in language generation models~\cite{sheng2019woman}.

Mitigating the existence of societal biases in data and models has been the topic of several studies. Some pieces of work propose the debiasing of word embeddings by identifying and removing gender subspace~\cite{ethayarajh2019understanding,bolukbasi2016man,kaneko2019gender,zhoa2018learning}. As pointed out by~\citet{gonen2019lipstick}, these methods successfully remove the \emph{explicit bias}, while the \emph{implicit bias}, examined through the ability of a classifier or a clustering algorithm to retrieve the gender of vectors after debiasing, still remains. Recently, \citet{lauscher2020general} propose a general framework for mitigating both forms of bias. 

Another approach to address bias, and related to this paper, is bias reduction in corpus via data augmentation. Counterfactual Data Augmentation (CDA) is a common method, which, in its basic form, extends a corpus by adding new sentences, achieved from swapping the indicative words of the concept of interest in the corpus. \citet{zhao2018gender} use CDA in the context of coreference resolution, \citet{lu2018gender} show the effectiveness of combination of CDA and embedding debiasing, \citet{zmigrod2019counterfactual} and later \citet{maudslay2019s} extend CDA to address bias in morphologically rich languages, as well as names. In this work, we use the basic CDA method to study the reaction of the bias measurement methods to the changes in the corpus.

\section{High-Order Bias Measurements}
We define bias as the discrepancy between the associations of a concept and its counterpart concept to a word. High-order bias measurement methods use vector similarity to quantify the associations. We define the concept $Z$ (and similarly its counterpart concept $Z'$) as a set, containing a group of representative words. In general, three approaches to high-order bias measurement are proposed in the literature, explained in what follows. 

\textbf{\biasPC:} In this method, first a matrix of directional vectors $\mD$ is created using a set of word pairs ${\sP_{Z,Z'}=\{(x,x')|x\in Z, x'\in Z'\}}$, such that ${\mD=\{\vv_x-\vv_{x'}|(x,x')\in\sP_{Z,Z'}\}}$, where $\vv_x$ is the vector of the word $x$. Using $\mD$, the first principle component of the directional vectors is then calculated, which we refer to as $\vv_{d}$.

\citet{bolukbasi2016man} define the bias of the word $w$ using the cosine similarity of $\vv_w$ and $\vv_{d}$. \citet{ethayarajh2019understanding} propose to normalize only $\vv_{d}$ to avoid the overestimation of the association degree, resulting in the following bias measurement:
\begin{equation}
\bias(w)=\frac{\vv_d\vv_w}{\Vert\vv_d\Vert}
\label{eq:background:biaspcdir}
\end{equation}
where $\bias(w)$ denotes the degree of bias of the word $w$, and its sign defines whether the word is biased towards the concept $Z$ or $Z'$. This definition is applicable to all the bias measurements, discussed through the paper.

\textbf{\biascentroid:} This method, used in several studies such as~\citet{garg2018word} and \citet{dev2019attenuating}, first defines the representative vector $\vv_Z$ as the mean of the embeddings of the representative words: 
\begin{equation}
\vv_Z=\sum_{x \in Z}{\frac{\vv_{x}}{\left|Z\right|}}
\label{eq:background:repvectorcent}
\end{equation}
The association of the concept $Z$ to the word $w$ is then defined using the cosine metric of $\vv_Z$ and $\vv_w$. Finally, the bias is calculated as follows:
\begin{equation}
\bias(w)=\text{cosine}(\vv_{Z}, \vv_w)-\text{cosine}(\vv_{Z'}, \vv_w)
\label{eq:background:repvector}
\end{equation}

\textbf{$\biasmeanSim$:} \citet{caliskan2017semantics} introduce Word Embedding Association Test (WEAT), a statistical test to examine the existence of bias using vector similarity. \citet{ethayarajh2019understanding} criticizes WEAT, showing that the conclusion of the test can be manipulated by swapping gender-related concept words. We therefore only study the method used by \citet{caliskan2017semantics} to measure the associations of words to concepts. This method calculates the average of the cosine similarities between the vector of the target word and the vectors of the concept words. We refer to this method as $\biasmeanSim$, formulated as follows:
\begin{equation}
\biasmeanSim(w,Z)=
\frac{1}{\left|Z\right|}\sum_{x \in Z}{\text{cosine}(\vv_x,\vv_w)}
\label{eq:background:mean}
\end{equation}
The bias using $\biasmeanSim$ is calculated as follows:
\begin{equation*}
\bias(w)=\biasmeanSim(w,Z)-\biasmeanSim(w,Z')
\label{eq:background:biasmeanSim}
\end{equation*}

\section{Novel Bias Measurement}
As discussed in Introduction, the first-order bias measurement requires the estimation of co-occurrence relations, which we provide using explicit representations. A well-known method to create explicit representations is based on the Point Mutual Information (PMI) metric. The PMI representation uses the count-based probabilities, where the co-occurrence relation between a word and a context-word in the PMI representation is calculated by ${\log \left(p(w,c)/p(w)p(c)\right)}$. Positive PMI (PPMI) is a commonly-used variation, where negative values are replaced with zero. \citet{levy2014neural} draw the relation between word2vec SkipGram (SG) embeddings and PMI representations, showing that SG can be seen as a factorization of the PMI matrix shifted by $\log k$. Based on this idea, they propose the Shifted Positive PMI (SPPMI) representation by subtracting $\log k$ from PMI vector representations and setting the negative values to zero.

In the following of this section, we first explain our approach to creating the explicit variations of the SG and GloVe vectors, referred to as \emph{explicit Skip-Gram (\hisg)} and \emph{explicit GloVe (\higlove)}. Our approach reconstructs explicit representations from embedding vectors, and is related to previous studies such as \citet{ethayarajh2019towards}, and \citet{levy2014neural}. We then describe our first-order bias measurement method, defined based on any explicit vector.

\subsection{Smoothed Explicit Representations}

\subsubsection{explicit Skip-Gram (\hisg)}
The original SG model consists of two parameter matrices: word ($\mV$) and context ($\mU$) matrices, both of size $\left|\sV\right| \times d$, where $\sV$ is the set of  words in the collection and $d$ is the embedding dimensionality. Given the word $c$, appearing in a context of word $w$, the model calculates $p(y=1|w,c)=\sigma(\vv_w\vu_{c}^\top)$, where $\vv_w$ is the vector representation of $w$, $\vu_c$ context-vector of $c$, and $\sigma$ denotes the sigmoid function. The SG model is optimized by maximizing the difference between $p(y=1|w,c)$ and $p(y=1|w,\check{c})$ for $k$ negative samples $\check{c}$, randomly drawn from a noisy distribution $\mathcal{N}$. The $\mathcal{N}$ distribution is set to the unigram distribution of the corpus, while downsampled by the context distribution smoothing parameter. 



The $p(y=1|w,c)$ term in the SG model measures the probability that the co-occurrence of two words $w$ and $c$ comes from the genuine co-occurrence distribution, derived from the training corpus. The model uses this probability to learn the embedding vectors, by separating these genuine co-occurrence relations from the sampled negative ones. We therefore use this estimation of the co-occurrence relations to define the vectors of the \hisg representation, resulting to the following definition of \hisg vector:

\begin{equation}
e_{w:c}=\sigma(\vv_w\vu_{c}^\top), \quad \ve_{w}=\sigma(\vv_w\mU^\top)\in \sR^{\left|\sV\right|}
\label{eq:method:esg}
\end{equation}
where $e_{w:c}$ denotes the value of the corresponding dimension of the vector of $w$ to the context-word $c$, and $\ve_{w}$ in $\left|\sV\right|$ dimensions is the explicit variation of the SG vector of word $w$.\footnote{An alternative formulation of \hisg is to normalize it by dividing its values with the square root of the expectations of the co-occurrence relations for each word and context-word, namely $\mathop{{}\mathbb{E}}_{c'\sim \mathcal{N}}{\sigma(\vv_w\vu_{c'}^\top)}$ and $\mathop{{}\mathbb{E}}_{w'\sim \mathcal{N}}{\sigma(\vv_{w'}\vu_{c}^\top)}$. We study this variation in pilot experiments, observing similar results to the introduced variation. We therefore stay with the less complex formulation (Eq.~\ref{eq:method:esg}).}

We should note that the \hisg representation is considerably different from the shifted PMI representation~\cite{levy2014neural}: shifted PMI assumes very high embedding dimensions during the model training, while \hisg draws the co-occurrence relations after the model is trained on low-dimensional embeddings. In fact, as SG is an implicit factorization of shifted PMI~\cite{levy2014neural,ethayarajh2019towards}, \hisg provides a smoothed variation of shifted PMI.

\subsubsection{explicit GloVe (\higlove)} 
The GloVe model first defines an explicit matrix (size $\left|\sV\right| \times \left|\sV\right|$), where the corresponding co-occurrence value of each word and context-word is set to ${\log p(w|c)}$. This log probability is calculated based on the number of co-occurrences (denoted by $\#\langle,\rangle$), such that ${\log p(w|c)=\log\#\langle w,c\rangle - \log\#\langle\cdot,c\rangle}$. We refer to this sparse explicit matrix as \textbf{\initglove}.

The GloVe model then implicitly factorizes the \initglove matrix, achieving two low-dimensional matrices of size $\left|\sV\right| \times d$, as well as two bias vectors of size $\left|\sV\right|$, where one assign a bias value to each word, and the other to each context-word. Using the same notation as SG, the factorization is done such that the dot products of the vectors of the matrices $\mV$ and $\mU$ plus the corresponding bias values estimate the logarithm of the co-occurrence values, as defined in the following:
\begin{equation}
\vv_w\vu_{c}^\top+b_{w} +\tilde{b}_c\approx \log \#\langle w,c\rangle
\label{eq:background:glove}
\end{equation}
where $b_{w}$ and $\tilde{b}_c$ denote the bias value of word $w$ and context-word $c$, respectively. 

Similar to \hisg, the \higlove representation estimates the co-occurrence relations using the word and context vectors, after training the GloVe model. Considering Eq.~\ref{eq:background:glove}, we define the co-occurrence relations of \higlove as the dot product of the word and context vectors,\footnote{As in \hisg, \higlove does not need extra normalization, since the bias terms $b_{w}$ and $\tilde{b}_c$ in Eq.~\ref{eq:background:glove}, learned during training, act as normalizers to the co-occurrence estimation ${\log \#\langle w,c\rangle}$.} shown as follows:
\begin{equation}
e_{w:c}=\vv_w\vu_c^\top,\quad \ve_w=\vv_w\mU^\top \in \sR^{\left|\sV\right|}
\label{eq:method:eglove}
\end{equation}

The \higlove representation in fact reconstructs \initglove, providing a smoothed variation.

\subsection{First-Order Bias Measurement} 
The main difference between the first-order bias measurement method and the high-order approaches is in the estimation of the associations of a word to the concepts. We define our bias measurement method based on the $\biasmeanSim$ method, by replacing the cosine metric with co-occurrence relations, and therefore refer to it as $\biasmeanCo$. Given an explicit vector denoted as $\ve$, $\biasmeanCo$ is defined as the mean of the co-occurrence values of $w$ with the representative words $Z$, formulated as follows:
\begin{equation}
\biasmeanCo(w,Z)=
\frac{1}{\left|Z\right|}\sum_{c \in Z}{e_{w:c}}
\label{eq:method:weam1st}
\end{equation}

The bias toward $Z$ is then defined as the differences between the associations of $w$ to $Z$ and $Z'$:
\begin{equation*}
\bias(w)=\biasmeanCo(w,Z)-\biasmeanCo(w,Z')
\label{eq:method:cbias}
\end{equation*}
As shown, $\biasmeanCo$ only considers the context-words related to the $Z$ and $Z'$ concepts. This avoids the influence of other non-relevant concepts as in the high-order bias measurements. 

Let us review the required calculations for $\biasmeanCo(w,Z)$ when using the introduced smoothed explicit representations, namely \hisg or \higlove. In this setting, the main computation of the bias measure is the dot products of the vector of $w$ to the context-vectors of the words in $Z$. We should note that the computational complexity of this calculation is the same as the one of $\biasmeanSim(w,Z)$ on SG/GloVe. In fact, considering the dot product of two embedding vectors as computation unit, the complexity of both bias measurement methods is $\mathcal{O}(\left|Z\right|)$. 

Finally, a practical consideration in calculating \hisg and \higlove is that this computation requires the context vectors in addition to word vectors. These context vectors are commonly stored in the libraries used for SG and GloVe embeddings alongside the word vectors, mainly for the purpose of continuous training. In \hisg/\higlove, these context vectors are exploited to estimate smoothed first-order relations.

\section{Gender Bias Experiment Design}
In this section, we explain the design of our experiment and the resources. Our source code together with all resources, including the lists of occupational and gender-representative words are publicly available.\footnote{\url{https://github.com/navid-rekabsaz/SmoothedFirstOrderBias}}.
 
\emph{Word Representations.} We conduct our experiments on the PMI, PPMI, SPPMI, SG, eSG, Glove, and eGlove representations. In addition, we create the low-dimensional vectors of the the PMI-based representations using Singular Value Decomposition (SVD), referred to as $\text{PMI-SVD}$, $\text{PPMI-SVD}$, and $\text{SPPMI-SVD}$. These word representation models are created on the English Wikipedia corpus of August 2017. We project all characters to lower case, and remove numbers and punctuation marks. For all models, we use the window size of 5, and filter the words with frequencies lower than 200, resulting in 197,549 unique words. The number of dimensions of the embeddings are set to 300. The rest of the parameters are set using the default parameter setting of the word2vec Skip-Gram model in the Gensim library~\cite{rehurek2010software}, and the GloVe model in the provided tool by its authors. As suggested by~\cite{levy2015improving}, we apply subsampling and context distribution smoothing on all PMI-based models with the same parameter values as the SG model. 

\emph{Gender-Representative Words.} In all bias measurements, the concepts $Z$ and $Z'$ are assumed as female and male, respectively. Therefore, a positive bias value indicates the inclination towards female, and a negative one towards male. The concepts are defined using two sets, each with 28 words, containing words like \emph{she}, \emph{her}, \emph{woman} for the female and \emph{he}, \emph{his}, \emph{man} for the male concept. These sets are taken from previous studies~\cite{bolukbasi2016man,garg2018word}. From the same sets, we form the gender pairs, used in the CDA method, and listed in Supplemental Material.

\emph{Occupational Words.} We provide a list of 496 occupations, among which 17 are female-specific (e.g.\ \emph{congresswoman}), 9 male-specific (e.g.\ \emph{congressman}), and the rest are gender-neutral (e.g.\ \emph{nurse} and \emph{dancer}). The set and assigned genders are listed in Supplemental Material.


\begin{table*}[h!]
\begin{center}
\centering
\begin{tabular}{l l l c c c c}
\toprule
\multirow{2}{*}{Order} &\multirow{2}{*}{Representation} & \multirow{2}{*}{Method} &\multicolumn{2}{c}{Labor Data} &\multicolumn{2}{c}{Census Data}  \\
& & & Spearman $\rho$ & Pearson's $r$ & Spearman $\rho$ & Pearson's $r$ \\\midrule

\multirow{6}{*}{High-Order} & \multirow{3}{*}{PMI} & $\biasPC$ & 0.28 & 0.07 & 0.18 & 0.02\\
& & $\biascentroid$ & 0.14 & 0.21 & 0.35 & 0.40\\
& & $\biasmeanSim$ & 0.33 & 0.24 & 0.27 & 0.19\\\cdashlinelr{3-7}
& \multirow{3}{*}{PMI-SVD} & $\biasPC$ & 0.05 & 0.07 & 0.00 & 0.00\\
& & $\biascentroid$ & 0.41 & 0.47 & 0.46 & 0.53\\
& & $\biasmeanSim$ & 0.41 & 0.49 & 0.49 & 0.56\\\cdashlinelr{1-7}
First-Order & PMI & $\biasmeanCo$ & \textbf{0.53} & \textbf{0.51} & \textbf{0.57} & \textbf{0.62}\\\midrule

\multirow{6}{*}{High-Order} & \multirow{3}{*}{PPMI} & $\biasPC$ & 0.45 & 0.49 & 0.39 & 0.47\\
& & $\biascentroid$ & 0.43 & 0.46 & 0.45 & 0.50\\
& & $\biasmeanSim$ & 0.43 & 0.46 & 0.45 & 0.52\\\cdashlinelr{3-7}
& \multirow{3}{*}{PPMI-SVD} & $\biasPC$ & 0.05 & 0.07 & 0.00 & 0.00\\
& & $\biascentroid$ & 0.41 & 0.47 & 0.46 & 0.53\\
& & $\biasmeanSim$ & 0.41 & 0.49 & 0.49 & 0.56\\ \cdashlinelr{1-7}
First-Order & PPMI & $\biasmeanCo$ & \textbf{0.59} & \textbf{0.58} & \textbf{0.64} & \textbf{0.64}\\\midrule

\multirow{6}{*}{High-Order} & \multirow{3}{*}{SPPMI} & $\biasPC$ & 0.26 & 0.37 & 0.26 & 0.28\\
& & $\biascentroid$ & 0.39 & 0.45 & 0.45 & \textbf{0.48}\\ 
& & $\biasmeanSim$ & 0.32 & 0.40 & 0.44 & \textbf{0.48}\\\cdashlinelr{3-7}
& \multirow{3}{*}{SPPMI-SVD} & $\biasPC$ & 0.17 & 0.29 & 0.11 & 0.03\\
& & $\biascentroid$ & 0.28 & 0.35 & 0.39 & 0.43\\ 
& & $\biasmeanSim$ & 0.26 & 0.38 & 0.36 & 0.46\\\cdashlinelr{1-7}
First-Order & SPPMI & $\biasmeanCo$ & \textbf{0.57} & \textbf{0.49} & \textbf{0.52} & \textbf{0.48}\\\midrule

\multirow{3}{*}{High-Order} & \multirow{3}{*}{GloVe} & $\biasPC$ & 0.53 & 0.56 & 0.34 & 0.46\\
& & $\biascentroid$ & 0.58 & 0.60 & 0.39 & 0.51 \\
& & $\biasmeanSim$ & \textbf{0.60} & \textbf{0.60} & 0.39 & 0.51 \\\cdashlinelr{1-7}
\multirow{2}{*}{First-Order} & \initglove & \multirow{2}{*}{$\biasmeanCo$} & 0.38 & 0.42 & 0.40 & 0.51\\
& \higlove & & 0.56 & 0.57 & \textbf{0.42} & \textbf{0.52}\\\midrule

\multirow{3}{*}{High-Order} & \multirow{3}{*}{SG} & $\biasPC$ & 0.50 & 0.54 & 0.58 & 0.64\\
& & $\biascentroid$ & 0.55 & 0.57 & 0.60 & 0.65 \\
& & $\biasmeanSim$ & 0.55 & 0.57 & 0.59 & 0.65\\\cdashlinelr{1-7}
First-Order & \hisg & $\biasmeanCo$ & \underline{\textbf{0.66}} & \underline{\textbf{0.61}} & \underline{\textbf{0.67}} & \underline{\textbf{0.70}}\\

\bottomrule
\end{tabular}
\caption{Spearman $\rho$ and Pearson's $r$ correlation results of the gender bias values, calculated with word representations, to the statistics of the portion of women in occupations}
\label{tbl:genderbias:correlation} 
\end{center}
\end{table*}

\emph{Job market statistics.} We study the correlation of the calculated gender bias values with the statistics of the gender bias of a set of occupations, obtained from the U.S. job market. These statistics are provided by two collections, where in each the bias for an occupation is the percent of people in the occupation who are reported as female (e.g. 90\% of nurses are women). The first collection uses the data provided by~\citet{zhao2018gender}. The collection contains the statistics of 40 occupations, gathered from the U.S. Department of Labor. We refer to the collection as Labor Data. The second is provided by~\citet{garg2018word} using the U.S. census data. From the provided data, we use the gender bias statistics of the year 2015 -- the most recent year in the collection, resulting in a list of 96 occupations. We refer to this as Census Data. We use Spearman $\rho$ and Pearson's $r$ correlations.


\section{Experiments and Results}
In this section, we first present the results of correlation studies with job market statistics, followed by analysing the results of the high-order bias measurement and discussing the characteristics of the two approaches. We then visualize and compare the gender bias of occupational words when measured with the first-order versus high-order approach, and finally study the reaction of the bias measurement approaches to bias reduction in the corpus.


\subsection{Correlation Results with Job Market Statistics} 
We calculate the gender bias of the occupations in the Labor Data and Census Data collections, using the high-order bias measurements (\biasPC, \biascentroid, and $\biasmeanSim$) on both low-dimensional and explicit representations, and the first-order bias measurement ($\biasmeanCo$) on explicit vectors. 

Table~\ref{tbl:genderbias:correlation} (next page) shows the results of the correlation between the calculated gender bias, and the gender bias statistics, provided by the Labor Data and Census Data collections. Each section of the table is assigned to a family of the models, namely PMI, PPMI, SPPMI, GloVe, and SG. The highest correlation results of each section are shown in bold, and the highest overall correlations are indicated with underlines.

In all families of representations, the first-order bias measurement shows higher correlations than the high-order methods, on both collections and correlation metrics (with the exception of GloVe on Labor Data). Overall, \hisg with $\biasmeanCo$ shows the highest correlations across the combinations. Interestingly, PPMI, as a simple count-based method, when combined with $\biasmeanCo$ shows higher correlations in comparison with any high-order bias measurement combined with SG or GloVe, on Census Data collection. 

The results indicate that the calculated bias values using the first-order method more closely follows the distributions of the gender bias indicators for occupations. This emphasizes the importance of exploiting the first-order bias measurement approach, as an alternative to the high-order approaches. The results of these experiments are particularly applicable to the studies, which explore the correlations of the text-based quantities with various external indicators (such as job market statistics)~\citet{caliskan2017semantics,garg2018word}, as in these work the quantities are only calculated based on the high-order bias measurements.

Comparing the correlation results of high-order methods, overall, $\biasmeanSim$ and \biascentroid show similar results, which are slightly higher than the ones of \biasPC. Considering the results of this study, in the following, we focus on analyzing the \hisg and SG representations, and $\biasmeanSim$ among the three high-order measurements.

\subsection{Analysis and Discussion}
In this section, we first diagnose the results of the high-order bias measurement, and then discus and compare the characteristics of the two bias measurement approaches. 

We start by exploring which context-words in practice contribute the most to the calculated gender bias of the occupational words with $\biasmeanSim$. To this end, we examine the measured gender bias of the occupational words using $\biasmeanSim$, applied on the \hisg vectors (instead of the SG vectors). We use \hisg since its explicit vectors enable the diagnosis of the results, particularly by looking at the context-words with the highest contributions.

\begin{table}[t]
\begin{center}
\centering
\begin{tabular}{L{7.5cm}}
\toprule

\textbf{manicurist:} businesswoman, \underline{nurse}, Filipina, seamstress, matron\\
\textbf{midwife:} \underline{midwife}, \underline{nurse}, \underline{feminist}, matron, \underline{suffragist}\\
\textbf{nurse:} \underline{midwife}, \underline{nurse}, matron, \underline{nursing}, Filipina\\
\textbf{socialite:} businesswoman, Filipina, \underline{suffragist}, \underline{feminist}, hostess\\
\textbf{housekeeper:} matron, \underline{midwife}, \underline{nurse}, maid, governess\\\midrule

\textbf{captain:} \underline{commanded}, \underline{capt}, \underline{quartermaster}, \underline{enlisted}, Hugh\\
\textbf{colonel:} \underline{commanded}, Hugh, Ernest, \underline{guards}, \underline{quartermaster}\\
\textbf{mechanician:} \underline{apprenticed}, Cyril, Ernest, Messrs, \underline{surveyor}\\
\textbf{lieutenant:} \underline{commanded}, Ernest, Hugh, \underline{enlisted}, \underline{quartermaster} \\
\textbf{engineer:} Jagmal, \underline{surveyor}, \underline{apprenticed}, \underline{draughtsman}, \underline{engineer} \\

\bottomrule
\end{tabular}
\caption{Context-words with the highest effects on the calculated gender bias with the $\biasmeanSim$ method. Gender-neutral context-words are shown with underlines.}
\label{tbl:genderbias:occpations_topdims} 
\end{center}
\end{table}

To diagnose the measured bias of $\biasmeanSim$, we first create the explicit variations of each female- and male-representative word vector, denoted as $\ve_z$ and $\ve_{z'}$ (see section Smoothed Explicit Representations). Given the occupation $w$ with explicit vector $\ve_w$, we calculate the high-order bias with $\biasmeanSim$, and provide its element-wise results by removing the summation of the cosine function, formulated as follows:
\begin{equation*}
\ve_{\bias}=
\frac{1}{\left|Z\right|}\sum_{z \in Z}{\frac{\ve_{z}\odot\ve_w}{\Vert\ve_{z}\Vert\Vert\ve_w\Vert}}-\frac{1}{\left|Z'\right|}\sum_{z' \in Z'}{\frac{\ve_{z'}\odot\ve_w}{\Vert\ve_{z'}\Vert\Vert\ve_w\Vert}}
\end{equation*}
where $\odot$ denotes the element-wise product, and $\ve_{\bias}\in \sR^{\left|\sV\right|}$ is the gender bias results of $w$ for all context-words, each correspond to one dimension.

To select a representative set of words for the analysis, we first calculate the gender bias of all target words using $\biasmeanSim$ on SG vectors. We then select the top 5 gender-neutral occupations with the highest bias towards female, and the same towards male. These top female and male words are shown in the top and bottom part of Table~\ref{tbl:genderbias:occpations_topdims}, respectively. For each word, we show the top 5 context-words with the largest influence on the results of the gender bias measurement, namely the highest absolute values of the dimensions in $\ve_{\bias}$.

In these representative samples, we can observe the existence of various groups of context-words: several gender-neutral context-words like \emph{nurse}, \emph{commanded}, and \emph{apprenticed} shown with underlines; several gender-specific words, which also contain other concepts such as an occupation like \emph{matron} and \emph{actress}; and finally person names such as \emph{Ernest} and \emph{Cyril}.


As discussed in Introduction, some of these context-words are arguably irrelevant to what we aim to characterize, namely the extent to which a word is perceived as female/male. For instance, the association of \emph{manicurist} with female is influenced by context-words such as \emph{nurse} and \emph{businesswoman}, or in the association of \emph{captain} with male is affected by \emph{commanded} and \emph{enlisted}. This issue is avoided in $\biasmeanCo$, as the method calculates the associations by only considering a pre-defined set of highly-relevant context-words.

Finally, let us put forward the question of \emph{which approach should be applied to bias measurement using text corpora?} Indeed answering this question, foremost, requires a clear definition of what is aimed to be captured as bias. Our study proposes $\biasmeanCo$, as a complementary to the existing high-order approaches, which exploits smoothed first-order co-occurrence relations of the highly-relevant context-words. This characteristic makes $\biasmeanCo$ particularly appropriate for bias measurement scenarios, in which capturing the direct relations between the words and the concept of interest is favorable. However, we should note that, depending on the objective of bias measurement, utilizing second- or higher-order co-occurrences can be also highly important. For instance, if the aim of a study is to measure the stereotypical bias of \emph{girls} and \emph{boys} towards the color \emph{pink}, it might be favorable to take into account the second-order co-occurrences, for example the ones through \emph{cloth}. However, even in this scenario, we should still be aware that utilizing the discussed high-order bias measurements might yet not be an ideal approach, as these methods are still affected by other non-relevant second- or higher-order context-words. An optimal bias measurement method for such a scenario still remains an open question, and studying it is a potential direction for future work. 


\begin{figure*}[tb]
\centering
\subfloat[High-order measurement ($\biasmeanSim$)]{\includegraphics[width=0.45\textwidth]{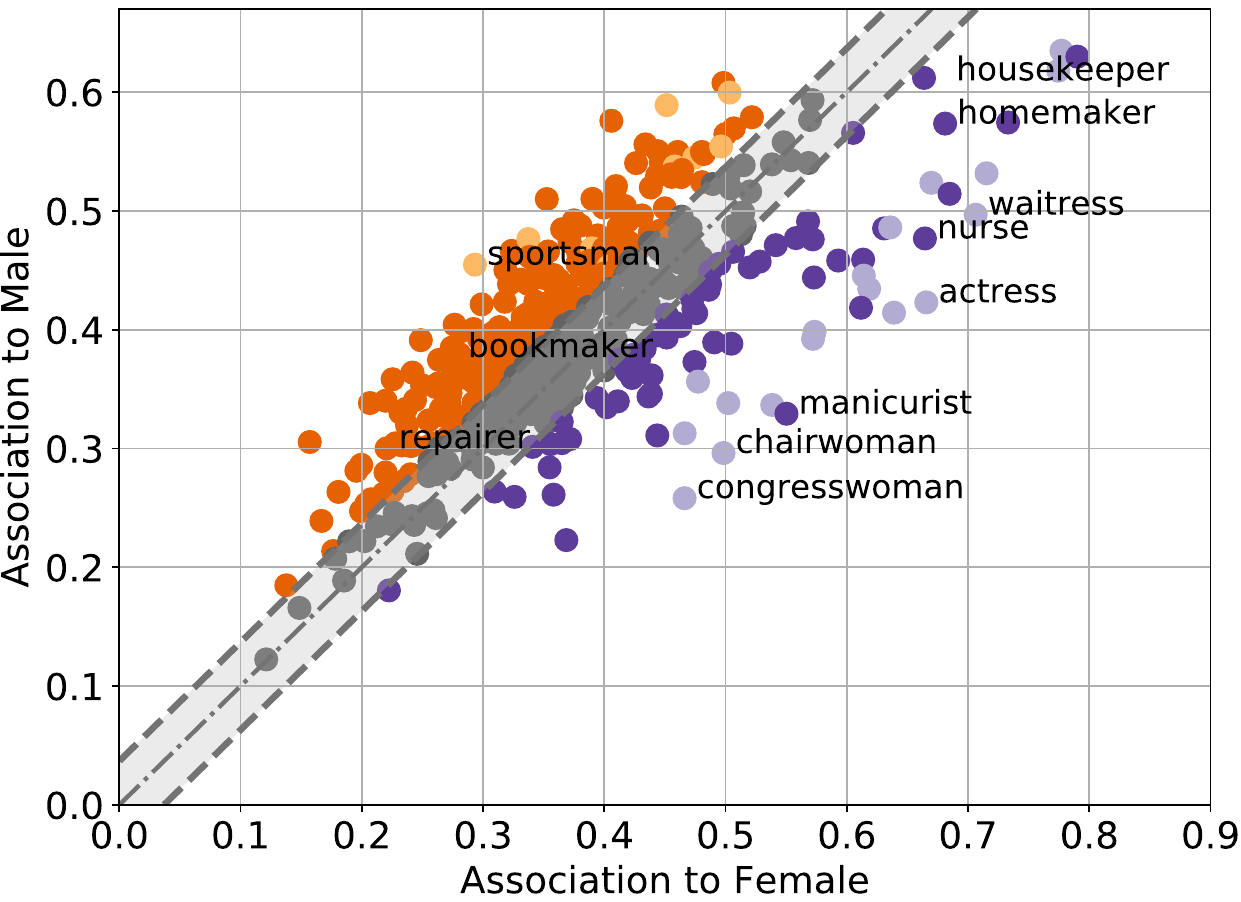}\label{fig:genderbias:genderbias_a}}
\qquad
\subfloat[First-order measurement ($\biasmeanCo$)]{\includegraphics[width=0.45\textwidth]{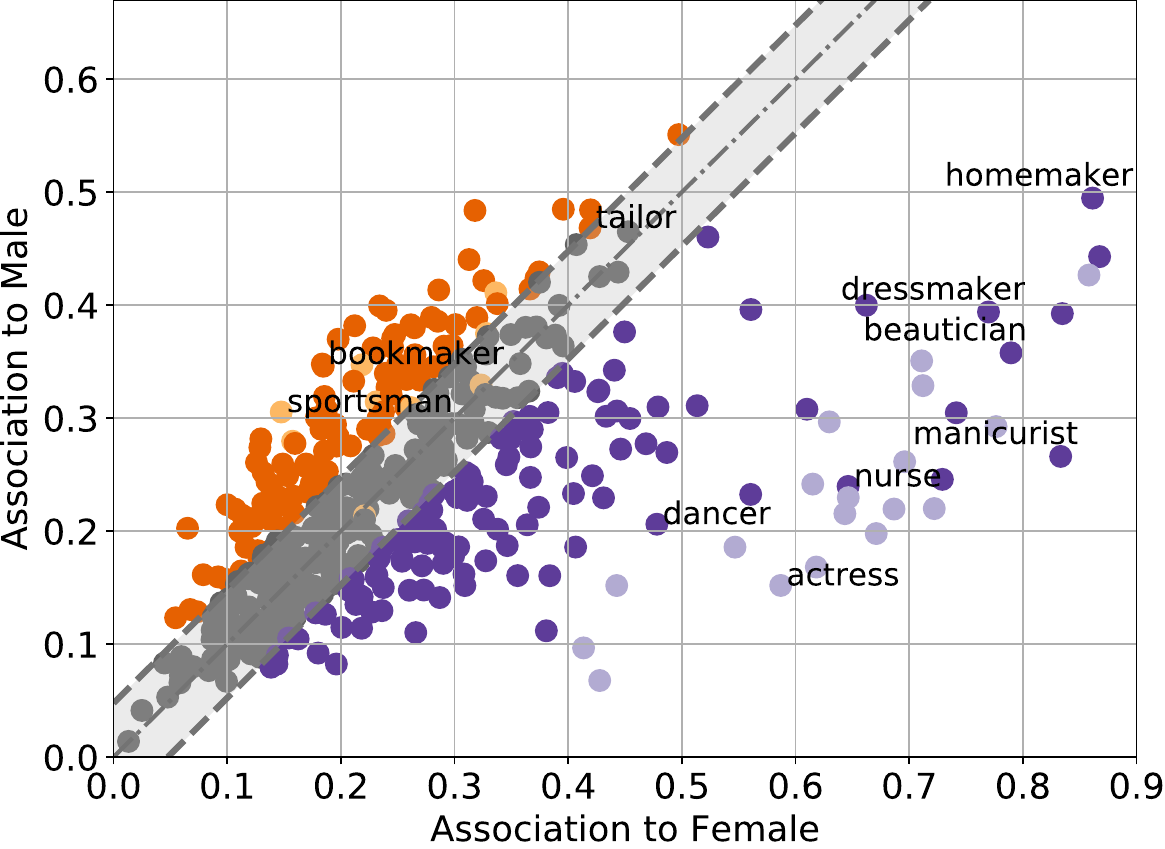}\label{fig:genderbias:genderbias_b}}
\caption{The associations of the occupations to the female and male concepts, indicating their inclinations towards the genders. The ones in the gray area are considered as unbiased. The occupations, inclined towards female and male are shown in purple and orange, respectively, where among them the gender-neutral ones, namely the occupations with stereotyped biases, have darker colors.}
\label{fig:genderbias:genderbias}
\end{figure*}

\subsection{Visualization of Gender Bias Results}
In this section, we continue our analyses by visualizing the gender bias of all occupations, measured using $\biasmeanSim$ on SG, and $\biasmeanCo$ on \hisg.

The results of the high- and first-order methods are shown in Figures~\ref{fig:genderbias:genderbias_a} and~\ref{fig:genderbias:genderbias_b}, respectively. To make the results visually comparable, we apply Min-Max normalization to the measured associations of each approach, where the min/max values are calculated over the measured associations of all words in vocabulary to male and female. 

In both bias measurement methods, we are interested in distinguishing between the words with significantly higher bias values and any random word with low bias values. To do this, we define a threshold for each plot, below which the words are considered as unbiased. To find the thresholds for distinguishing between the words with significantly higher bias values and any random word with low bias values, since the number of biased words to a concept are limited, we assume that there is a high probability that any randomly sampled word is unbiased. We therefore define the threshold for each bias measurement method as the mean of the absolute bias values of all words in vocabulary. These thresholds for $\biasmeanSim$ with SG and $\biasmeanCo$ with \hisg are $0.036$ and $0.047$, respectively.

In each plot, the area where the distance from the diagonal is less than the corresponding threshold value, is referred to as the unbiased area, and shown in gray. The occupations located in the unbiased areas are considered as unbiased (no significant stereotypical bias is observed). 


A gender-neutral occupation is considered to be stereotypically biased to either female or male, when it is inclined towards the female/male associations, namely when it is located below/above the unbiased area, shown in dark purple and dark orange, respectively. The plot also depicts the gender-specific occupations (e.g.\ \emph{actress}, \emph{sportman}), while we should consider that these occupations are not stereotypically biased, but are expected to be inclined towards their respective genders. We show these gender-specific occupations in light colours, namely light purple for female, and light orange for male. As expected, all gender-specific occupation words are on the correct side of the diagonal for both measures.  



Both figures show the existence of significant gender bias in several occupations. However, Figure~\ref{fig:genderbias:genderbias_a} and~\ref{fig:genderbias:genderbias_b} provide considerably different perspectives on the extents of the associations of the occupations to genders. In particular, the $\biasmeanCo$ method shows relatively larger degrees of bias towards female, specially for some gender-neutral occupations such as \emph{nurse}, and \emph{housekeeper}.

\begin{figure}[t]
\centering
\subfloat[$\biasmeanSim$]{\includegraphics[width=0.24\textwidth]{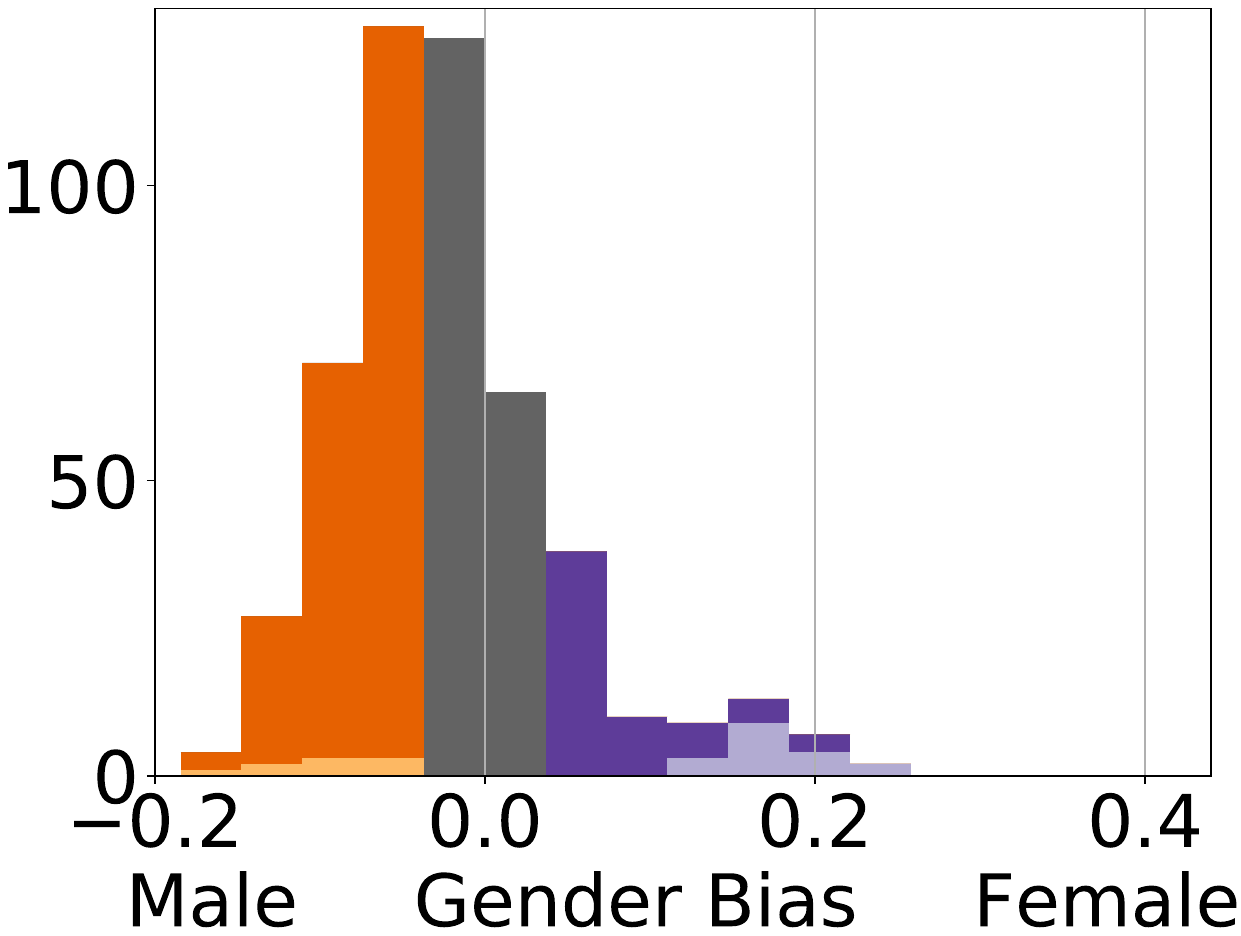}\label{fig:genderbias:genderbias_hist_a}}
\subfloat[$\biasmeanCo$]{\includegraphics[width=0.24\textwidth]{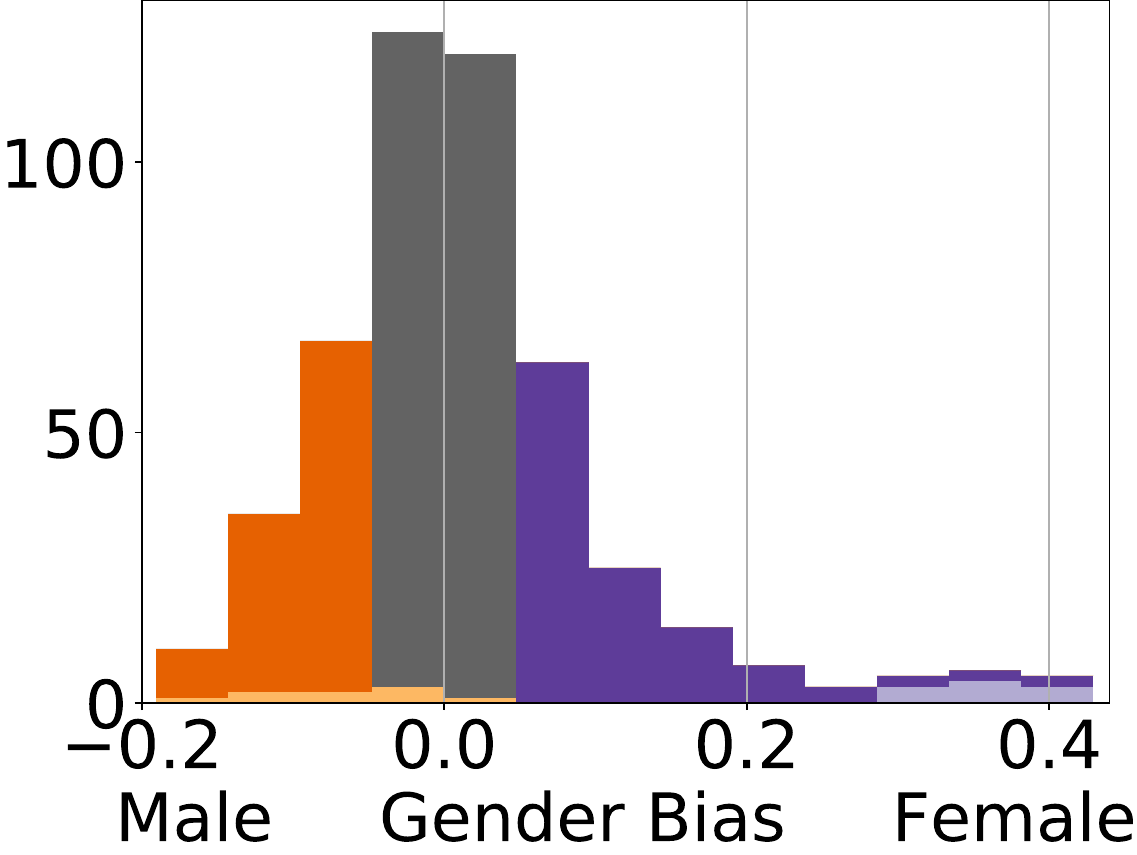}\label{fig:genderbias:genderbias_hist_b}}
\caption{Histograms of the gender bias of the occupations, measured using $\biasmeanSim$ and $\biasmeanCo$. }
\label{fig:genderbias:genderbias_hist}
\end{figure}

To have a better view on the distribution of the bias values, Figure~\ref{fig:genderbias:genderbias_hist} shows the histogram of the occupations over the range of the bias values, measured with $\biasmeanSim$ and $\biasmeanCo$. Similar to Figure~\ref{fig:genderbias:genderbias}, the gray color shows the number of unbiased occupations, and the purple and orange colors indicate the number of biased ones towards female and male in each bin, among which the gender-specific ones are shown with light colors.

In both measurement methods, a larger number of occupations are biased to male. However, the first-order bias measurement captures a larger degree of bias towards female, suggesting the existence of a more severe degree of female bias, previously undetected by the high-order approach.

\begin{figure}[t]
\centering\includegraphics[width=0.4\textwidth]{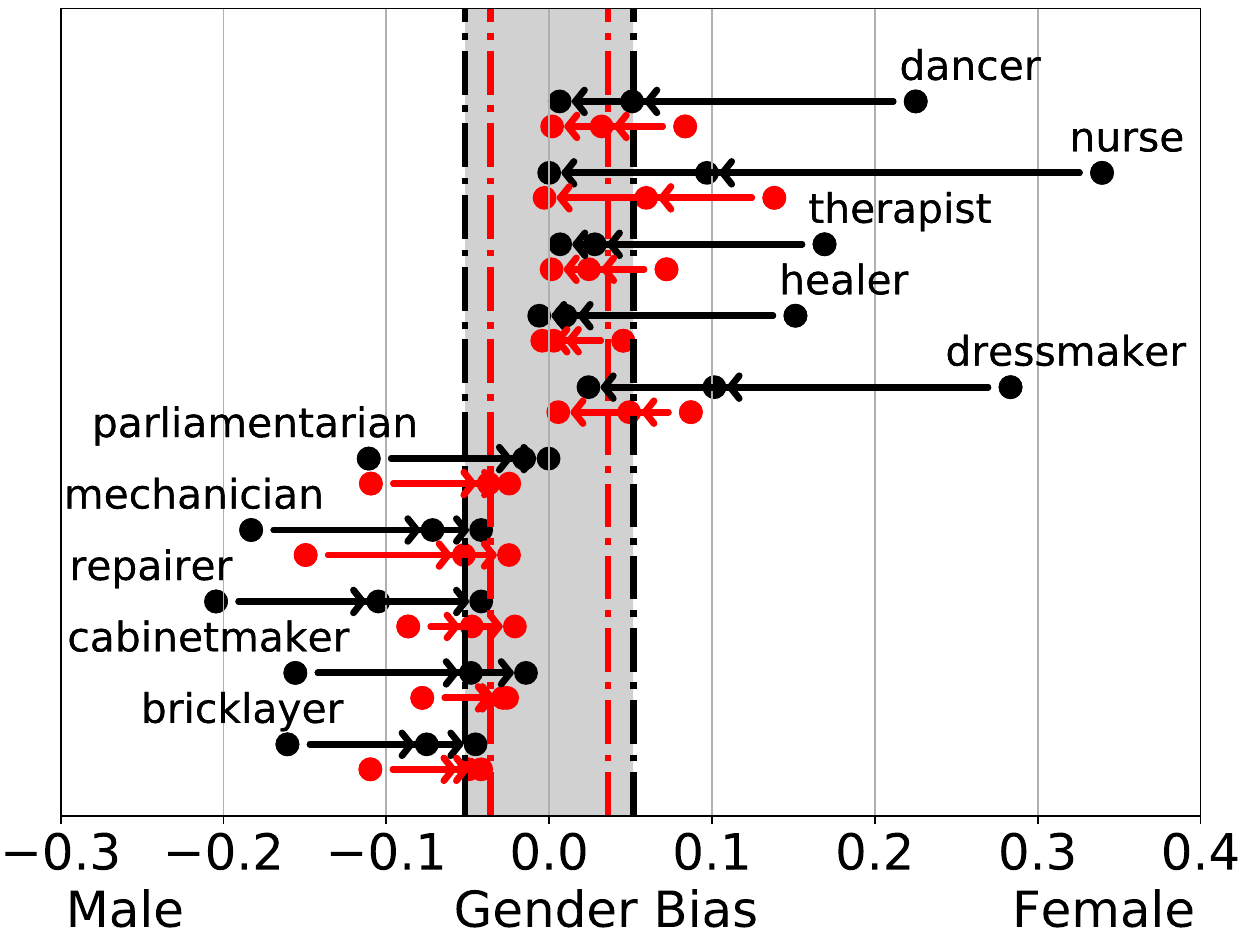}
\caption{Changes in the gender bias values, measured on the original corpus and two augmented ones with the CDA method. The black and red arrows indicate the results of $\biasmeanCo$ and $\biasmeanSim$, respectively.}
\label{fig:genderbias:swapped}
\end{figure}

\subsection{Sensitivity to Changes in Corpus}
How do the bias measurement methods react to (hypothetical) future changes, especially when the provided corpora move towards more balanced representations of genders? To explore this question, we use the basic form of the CDA method (see Related Work) to expand the corpus with synthetic augmented data. Using the list of gender words pairs, for every sentence in the corpus, we swap each female word with its counter-part male word, and vice versa. We create two new augmented corpora: the first one expands the original corpus with only one half of the new sentences (selected randomly), while the second one adds all the new sentences to the original corpus. 

We create the SG models, and consecutively the \hisg representations, based on these two new corpora. We then measure gender bias of the corpora using SG with $\biasmeanSim$, and \hisg vectors with $\biasmeanCo$. Figure~\ref{fig:genderbias:swapped} shows the results of the changes in the bias values of 5 randomly-selected gender-neutral occupations for each gender, which are perceived as biased in the original corpus by both bias measurements. Each arrow shows the direction of the changes, where the starting point indicates the measured bias in the original corpus, and the middle and final points are calculated based on the first and second augmented corpora. The black and red arrows show the results of $\biasmeanCo$ and $\biasmeanSim$, respectively. The gray areas, bordered with the black and red vertical dashed lines show the unbiased areas of the results of $\biasmeanCo$ and $\biasmeanSim$, respectively. As mentioned in previous subsection, since in each method the values are normalized over all words, the bias values across the methods are reasonably comparable.

As shown, in both methods the gender bias of all occupations consistently decrease (move towards zero). The gender bias values measured on the second augmented corpus, for all the occupations, reach the corresponding unbiased area of each method, indicating that these occupations in this corpus are considered as unbiased. Comparing the bias measurement methods, for these occupations, the $\biasmeanCo$ approach indicates the existence of a higher degree of bias in the initial corpus in comparison with $\biasmeanSim$, and also demonstrates a larger degree of bias reduction in each step. The aggregated degrees of changes over all gender-neutral occupations are reported Table~\ref{tbl:biasreduction}. As shown, in both steps the $\biasmeanCo$ demonstrate larger changes in comparison with $\biasmeanSim$.

These larger changes in the measured bias indicate the higher sensitivity of the $\biasmeanCo$ approach to the changes in the corpus in comparison with $\biasmeanSim$, which is a valuable characteristic of the first-order method particularly in the study of social changes using text data.

\begin{table}[t]
\centering
\begin{tabular}{l C{1.8cm} C{1.8cm}}
\toprule
Corpus Augmentation & $\biasmeanSim$ with SG & $\biasmeanCo$ with \hisg \\\midrule
First Step & 0.037 & \textbf{0.044} \\
Second Step & 0.012 & \textbf{0.016} \\
\bottomrule
\end{tabular}
\caption{Mean of the absolute values of bias changes in each step of corpus augmentation experiments, discussed}
\label{tbl:biasreduction} 
\end{table}


\section{Conclusion}
We highlight a potential issue of the commonly-used high-order bias measurement approaches, and propose a novel approach based on the smoothed first-order co-occurrence relations between words and their context words. To estimate these relations, we reconstruct explicit vectors from pre-trained word2vec Skip-Gram, and GloVe embeddings, proposing smoothed first-order bias measurement method. 

Let us visit again the question of \emph{``which approach should we use for measuring bias (or any societal phenomena) using text corpora?''} To answer this question, we recommend the reader to first contemplate: \emph{``what construct/quality is aimed to be captured as bias''}. As an example, measuring \emph{``to what extent different gender-neutral occupations are perceived differently''} -- as pursued in our experiments -- arguably looks for a different construct/realization of bias in comparison with measuring \emph{``what the stereotypical relation of color pink is towards girls and boys.''}. The smoothed first-order approach offers an alternative to high-order approaches, and particularly has the advantage of narrowing down the measurement to the effect of only a specific set of representative words, appearing in the vicinity of the word under study. 

To have a better understanding of the differences of these two approaches, our experiments examine and compare the measured gender bias of a set of occupational words, calculated on a Wikipedia text corpus. We observe that the gender bias calculated with the proposed first-order approach correlates higher with the gender-related statistics of the job market in comparison to the high-order methods. Our analyses provide showcases of the existence of non-relevant context-words when using high-order measurements, and suggests the existence of a more severe degree of female-bias regarding some jobs. While our experiments are conducted on gender bias, the introduced method is generic and can be extended to study other forms of societal biases such as related to race, age and ethnicity.

\appendix







\section{Supplemental Material}

\subsection{Gender Definitional Words}
\small
\paragraph{Female:} \emph{girl}, \emph{girls},  \emph{sister}, \emph{sisters}, \emph{mom}, \emph{moms}, \emph{mother}, \emph{mothers}, \emph{fiancee}, \emph{grandmother}, \emph{grandma}, \emph{granddaughter}, \emph{granddaughters}, \emph{she}, \emph{her}, \emph{herself}, \emph{hers}, \emph{gal}, \emph{gals}, \emph{female}, \emph{females}, \emph{woman}, \emph{women}, \emph{madam}, \emph{daughter}, \emph{daughters}, \emph{stepmother}, \emph{stepdaughter}

\paragraph{Male:} \emph{boy}, \emph{boys}, \emph{brother}, \emph{brothers}, \emph{dad}, \emph{dads}, \emph{father}, \emph{fathers}, \emph{fiance}, \emph{grandfather}, \emph{grandpa}, \emph{grandson}, \emph{grandsons}, \emph{he}, \emph{him}, \emph{himself}, \emph{his}, \emph{lad}, \emph{lads}, \emph{male}, \emph{males}, \emph{man}, \emph{men}, \emph{sir}, \emph{son}, \emph{sons}, \emph{stepfather}, \emph{stepson}

\paragraph{Pairs:} (\emph{boy, girl}), (\emph{boys, girls}), (\emph{brother, sister}), (\emph{brothers, sisters}), (\emph{dad, mom}), (\emph{dads, moms}), (\emph{father, mother}), (\emph{fathers, mothers}), (\emph{fiance, fiancee}), (\emph{grandfather, grandmother}), (\emph{grandpa, grandma}), (\emph{grandson, granddaughter}), (\emph{grandsons, granddaughters}), (\emph{he, she}), (\emph{him, her}), (\emph{himself, herself}), (\emph{his, hers}), (\emph{lad, gal}), (\emph{lads, gals}), (\emph{male, female}), (\emph{males, females}), (\emph{man, woman}), (\emph{men, women}), (\emph{sir, madam}), (\emph{son, daughter}), (\emph{sons, daughter}), (\emph{stepfather, stepmother}), (\emph{stepson, stepdaughter})

\subsection{Target Words}
\small
\paragraph{Female-specific occupations:} \emph{actress}, \emph{ballerina}, \emph{barmaid}, \emph{businesswoman}, \emph{chairwoman}, \emph{chambermaid}, \emph{congresswoman}, \emph{forewoman}, \emph{housewife}, \emph{landlady}, \emph{maid}, \emph{masseuse}, \emph{matron}, \emph{mistress}, \emph{policewoman}, \emph{sportswoman}, \emph{stewardess}, \emph{stuntwoman}, \emph{usherette}, \emph{waitress}

\paragraph{Male-specific occupations:} \emph{barman}, \emph{businessman}, \emph{foreman}, \emph{landlord}, \emph{policeman}, \emph{salesman}, \emph{serviceman}, \emph{sportsman}, \emph{waiter}

\paragraph{Gender-neutral occupations:} \emph{accountant}, \emph{actor}, \emph{administrator}, \emph{adventurer}, \emph{adviser}, \emph{advocate}, \emph{agent}, \emph{aide}, \emph{ambassador}, \emph{analyst}, \emph{animator}, \emph{announcer}, \emph{anthropologist}, \emph{apprentice}, \emph{archaeologist}, \emph{archeologist}, \emph{architect}, \emph{archivist}, \emph{artist}, \emph{artiste}, \emph{assassin}, \emph{assessor}, \emph{assistant}, \emph{astrologer}, \emph{astronaut}, \emph{astronomer}, \emph{athlete}, \emph{attendant}, \emph{attorney}, \emph{auctioneer}, \emph{auditor}, \emph{author}, \emph{bailiff}, \emph{baker}, \emph{ballplayer}, \emph{banker}, \emph{barber}, \emph{bargee}, \emph{baron}, \emph{barrister}, \emph{bartender}, \emph{basketmaker}, \emph{beautician}, \emph{beekeeper}, \emph{bibliographer}, \emph{biochemist}, \emph{biologist}, \emph{biotechnologist}, \emph{blacksmith}, \emph{boatman}, \emph{bodyguard}, \emph{boilerfitter}, \emph{boilermaker}, \emph{bookbinder}, \emph{bookkeeper}, \emph{bookmaker}, \emph{bootmaker}, \emph{boss}, \emph{botanist}, \emph{boxer}, \emph{breeder}, \emph{brewer}, \emph{bricklayer}, \emph{broadcaster}, \emph{broker}, \emph{bureaucrat}, \emph{butcher}, \emph{butler}, \emph{buyer}, \emph{cabbie}, \emph{cabinetmaker}, \emph{cameraman}, \emph{campaigner}, \emph{captain}, \emph{cardiologist}, \emph{caretaker}, \emph{carpenter}, \emph{cartographer}, \emph{cartoonist}, \emph{cashier}, \emph{cellist}, \emph{ceo}, \emph{ceramicist}, \emph{chairman}, \emph{chancellor}, \emph{chaplain}, \emph{character}, \emph{chef}, \emph{chemist}, \emph{chief}, \emph{chiropractor}, \emph{choreographer}, \emph{cinematographer}, \emph{citizen}, \emph{cleaner}, \emph{clergy}, \emph{cleric}, \emph{clerical}, \emph{clerk}, \emph{coach}, \emph{collector}, \emph{colonel}, \emph{columnist}, \emph{comedian}, \emph{comic}, \emph{commander}, \emph{commentator}, \emph{commissioner}, \emph{composer}, \emph{compositor}, \emph{concreter}, \emph{conductor}, \emph{confectioner}, \emph{confesses}, \emph{congressman}, \emph{conservationist}, \emph{conservator}, \emph{constable}, \emph{consultant}, \emph{cook}, \emph{cooper}, \emph{cop}, \emph{coremaker}, \emph{correspondent}, \emph{councilman}, \emph{councilor}, \emph{counsellor}, \emph{counselor}, \emph{courtier}, \emph{critic}, \emph{crooner}, \emph{croupier}, \emph{crusader}, \emph{crusher}, \emph{curator}, \emph{custodian}, \emph{cutler}, \emph{dancer}, \emph{dean}, \emph{decorator}, \emph{dentist}, \emph{deputy}, \emph{dermatologist}, \emph{designer}, \emph{detective}, \emph{developer}, \emph{dietician}, \emph{dietitian}, \emph{digger}, \emph{diplomat}, \emph{director}, \emph{dispatcher}, \emph{diver}, \emph{doctor}, \emph{doorkeeper}, \emph{draughtsman}, \emph{draughtsperson}, \emph{dresser}, \emph{dressmaker}, \emph{driller}, \emph{driver}, \emph{drummer}, \emph{drycleaner}, \emph{dyer}, \emph{ecologist}, \emph{economist}, \emph{editor}, \emph{educator}, \emph{electrician}, \emph{embroiderer}, \emph{employee}, \emph{engineer}, \emph{engraver}, \emph{entertainer}, \emph{entrepreneur}, \emph{environmentalist}, \emph{envoy}, \emph{epidemiologist}, \emph{ergonomist}, \emph{ethnographer}, \emph{evangelist}, \emph{expert}, \emph{farmer}, \emph{farrier}, \emph{filmmaker}, \emph{financier}, \emph{firebrand}, \emph{firefighter}, \emph{fisherman}, \emph{fitter}, \emph{footballer}, \emph{furrier}, \emph{gangster}, \emph{gardener}, \emph{geneticist}, \emph{geographer}, \emph{geologist}, \emph{geophysicist}, \emph{gilder}, \emph{glassmaker}, \emph{glazier}, \emph{goalkeeper}, \emph{goatherd}, \emph{goldsmith}, \emph{gravedigger}, \emph{grinder}, \emph{guard}, \emph{guide}, \emph{guitarist}, \emph{gunsmith}, \emph{hairdresser}, \emph{handler}, \emph{handyman}, \emph{hardener}, \emph{harpooner}, \emph{hatter}, \emph{headmaster}, \emph{healer}, \emph{herbalist}, \emph{historian}, \emph{homemaker}, \emph{hooker}, \emph{housekeeper}, \emph{hydrologist}, \emph{illustrator}, \emph{industrialist}, \emph{infielder}, \emph{inspector}, \emph{instructor}, \emph{insulator}, \emph{interpreter}, \emph{inventor}, \emph{investigator}, \emph{janitor}, \emph{jeweler}, \emph{jeweller}, \emph{joiner}, \emph{journalist}, \emph{judge}, \emph{jurist}, \emph{keeper}, \emph{knitter}, \emph{laborer}, \emph{labourer}, \emph{lacemaker}, \emph{lawmaker}, \emph{lawyer}, \emph{lecturer}, \emph{legislator}, \emph{librarian}, \emph{lieutenant}, \emph{lifeguard}, \emph{lithographer}, \emph{lumberjack}, \emph{lyricist}, \emph{machinist}, \emph{maestro}, \emph{magician}, \emph{magistrate}, \emph{maltster}, \emph{manager}, \emph{manicurist}, \emph{marksman}, \emph{marshal}, \emph{mason}, \emph{masseur}, \emph{master}, \emph{mathematician}, \emph{mechanic}, \emph{mechanician}, \emph{mediator}, \emph{medic}, \emph{melter}, \emph{merchandiser}, \emph{metallurgist}, \emph{metalworker}, \emph{meteorologist}, \emph{metrologist}, \emph{microbiologist}, \emph{midfielder}, \emph{midwife}, \emph{miller}, \emph{millwright}, \emph{miner}, \emph{minister}, \emph{missionary}, \emph{mobster}, \emph{model}, \emph{modeller}, \emph{mover}, \emph{musician}, \emph{musicologist}, \emph{nanny}, \emph{narrator}, \emph{naturalist}, \emph{negotiator}, \emph{neurologist}, \emph{neurosurgeon}, \emph{novelist}, \emph{nurse}, \emph{nutritionist}, \emph{observer}, \emph{obstetrician}, \emph{officer}, \emph{official}, \emph{operator}, \emph{optician}, \emph{optometrist}, \emph{organiser}, \emph{organist}, \emph{orthotist}, \emph{owner}, \emph{packer}, \emph{paediatrician}, \emph{painter}, \emph{palmists}, \emph{paperhanger}, \emph{paralegal}, \emph{paramedic}, \emph{parishioner}, \emph{parliamentarian}, \emph{pastor}, \emph{pathologist}, \emph{patrolman}, \emph{patternmaker}, \emph{paver}, \emph{pawnbroker}, \emph{pediatrician}, \emph{pedicurist}, \emph{performer}, \emph{pharmacist}, \emph{philanthropist}, \emph{philosopher}, \emph{photographer}, \emph{photojournalist}, \emph{physician}, \emph{physicist}, \emph{physiotherapist}, \emph{pianist}, \emph{pilot}, \emph{planner}, \emph{plasterer}, \emph{playwright}, \emph{plumber}, \emph{poet}, \emph{police}, \emph{politician}, \emph{pollster}, \emph{porter}, \emph{potter}, \emph{poulterer}, \emph{preacher}, \emph{president}, \emph{priest}, \emph{principal}, \emph{prisoner}, \emph{producer}, \emph{professor}, \emph{programmer}, \emph{projectionist}, \emph{promoter}, \emph{prompter}, \emph{proprietor}, \emph{prosecutor}, \emph{prosthetist}, \emph{protagonist}, \emph{protege}, \emph{protester}, \emph{provost}, \emph{psychiatrist}, \emph{psychologist}, \emph{psychotherapist}, \emph{publicist}, \emph{publisher}, \emph{pundit}, \emph{radiographer}, \emph{radiologist}, \emph{radiotherapist}, \emph{ranger}, \emph{realtor}, \emph{receptionist}, \emph{referee}, \emph{refiner}, \emph{registrar}, \emph{repairer}, \emph{reporter}, \emph{representative}, \emph{rescuer}, \emph{researcher}, \emph{restaurateur}, \emph{retoucher}, \emph{rigger}, \emph{roaster}, \emph{roofer}, \emph{sailor}, \emph{saint}, \emph{sales}, \emph{salesperson}, \emph{sausagemaker}, \emph{saxophonist}, \emph{scaffolder}, \emph{scholar}, \emph{scientist}, \emph{screenwriter}, \emph{scriptwriter}, \emph{sculptor}, \emph{seaman}, \emph{secretary}, \emph{senator}, \emph{sergeant}, \emph{servant}, \emph{setter}, \emph{sewer}, \emph{shepherd}, \emph{sheriff}, \emph{shoemaker}, \emph{shopkeeper}, \emph{shunter}, \emph{singer}, \emph{skipper}, \emph{smith}, \emph{socialite}, \emph{sociologist}, \emph{soldier}, \emph{solicitor}, \emph{soloist}, \emph{songwriter}, \emph{specialist}, \emph{spinner}, \emph{sportswriter}, \emph{staff}, \emph{statesman}, \emph{statistician}, \emph{steelworker}, \emph{steeplejack}, \emph{steward}, \emph{stockbroker}, \emph{stonecutter}, \emph{stonemason}, \emph{storekeeper}, \emph{strategist}, \emph{student}, \emph{stuntman}, \emph{stylist}, \emph{substitute}, \emph{superintendent}, \emph{supervisor}, \emph{surgeon}, \emph{surveyor}, \emph{sweep}, \emph{swimmer}, \emph{tailor}, \emph{tamer}, \emph{tanner}, \emph{tannery}, \emph{teacher}, \emph{technician}, \emph{technologist}, \emph{telecaster}, \emph{teller}, \emph{therapist}, \emph{tinsmith}, \emph{toolmaker}, \emph{tracklayer}, \emph{trader}, \emph{trainer}, \emph{trainman}, \emph{translator}, \emph{traveller}, \emph{treasurer}, \emph{trooper}, \emph{trucker}, \emph{trumpeter}, \emph{tuner}, \emph{turner}, \emph{tutor}, \emph{tycoon}, \emph{typesetter}, \emph{tyrefitter}, \emph{undersecretary}, \emph{understudy}, \emph{upholsterer}, \emph{usher}, \emph{valedictorian}, \emph{valuer}, \emph{varnisher}, \emph{vendor}, \emph{veterinarian}, \emph{viniculturist}, \emph{violinist}, \emph{vocalist}, \emph{warden}, \emph{warrior}, \emph{washer}, \emph{weaver}, \emph{weigher}, \emph{welder}, \emph{whaler}, \emph{wigmaker}, \emph{worker}, \emph{wrestler}, \emph{writer}, \emph{zookeeper}




\normalsize

\bibliography{references}

\end{document}